\pdfoutput=1

\documentclass[11pt]{article}

\usepackage{acl}
\usepackage{algorithm}
\usepackage{algpseudocode}
\usepackage{geometry}
\usepackage[utf8]{inputenc}
\usepackage{tcolorbox}
\usepackage{times}
\usepackage{latexsym}
\usepackage[T1]{fontenc}
\usepackage[utf8]{inputenc}
\usepackage{microtype}
\usepackage{inconsolata}
\usepackage{graphicx}
\usepackage{subcaption}
\usepackage{bm}
\usepackage{comment}
\usepackage{enumitem}
\usepackage{amsmath}
\usepackage{amssymb}
\usepackage{makecell}
\usepackage{tabularx}

\title{Which Retain Set Matters for LLM Unlearning?

A Case Study on Entity Unlearning}

\author{Hwan Chang \and Hwanhee Lee\thanks{Corresponding author.} \\
    Department of Artificial Intelligence, Chung-Ang University, Seoul, Korea\\
    \texttt{\{hwanchang, hwanheelee\}@cau.ac.kr}
}
\begin{document}
\maketitle
\begin{abstract}
Large language models (LLMs) risk retaining unauthorized or sensitive information from their training data, which raises privacy concerns. LLM unlearning seeks to mitigate these risks by selectively removing specified data while maintaining overall model performance. However, most existing work focuses on methods to achieve effective forgetting and does not provide a detailed analysis of the retain set, the portion of training data that is not targeted for removal. 
In this paper, we investigate the effects of unlearning on various subsets of the retain set through a case study on entity unlearning. We introduce the \textit{Syntactically Similar Neighbor Set}, a group of queries that share similar syntactic structures with the data targeted for removal, and show that this subset suffers the greatest performance drop during unlearning. Moreover, when used for regularization, this set not only preserves performance on syntactically similar queries but also delivers comparable or improved results across other data subsets. Our results highlight that syntactic similarity is a critical factor, potentially more so than domain or entity relationships, in achieving effective and practical LLM unlearning.
\end{abstract}

\section{Introduction}
\begin{figure}[!t]
\label{fig:introFig}
\includegraphics[width=\linewidth]{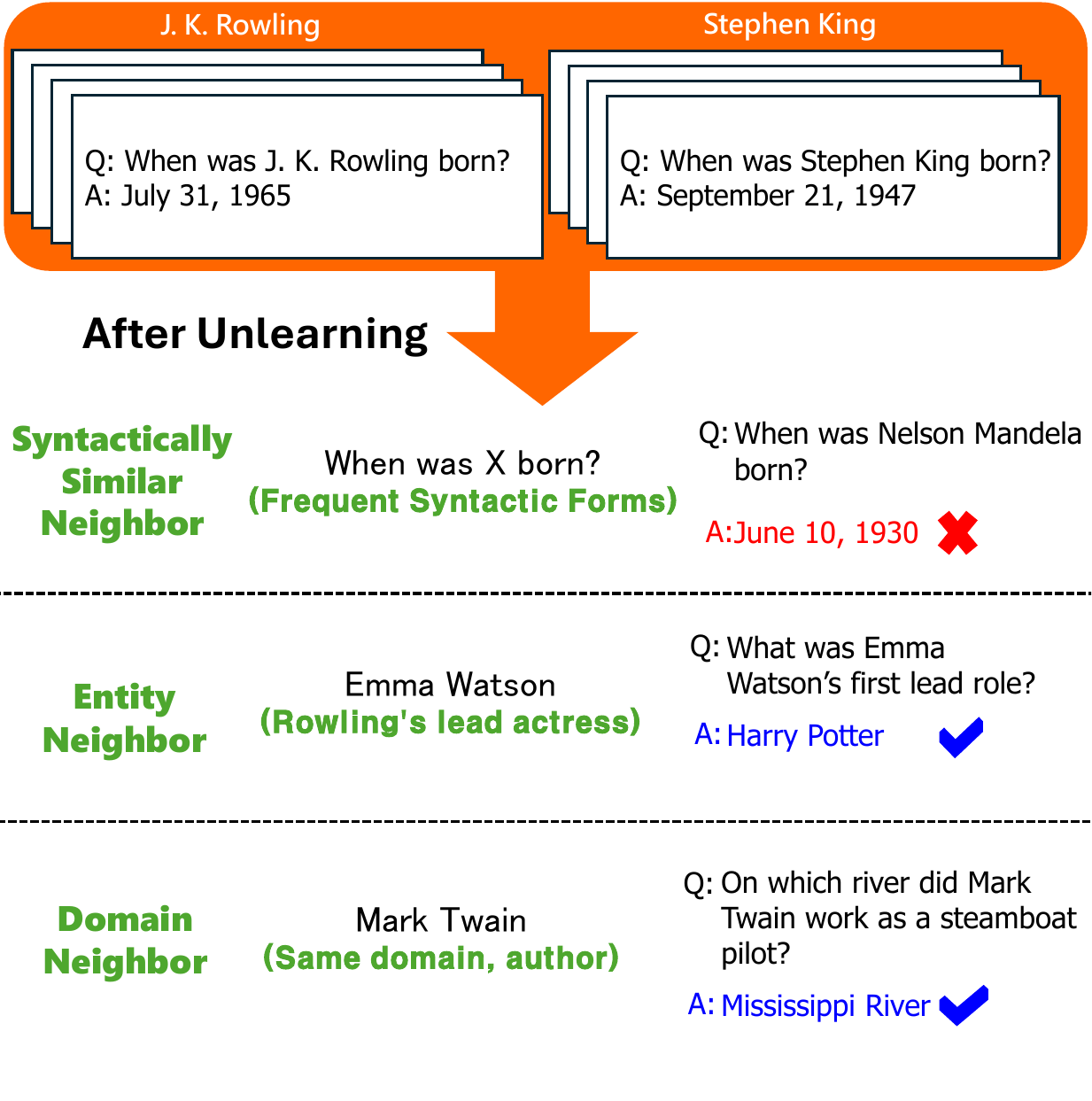}
\vspace{-10mm}
\caption{Impact of unlearning across different neighbor sets. Syntactically similar neighbors are most affected (in red). In contrast, entity and domain neighbors retain correct knowledge (in blue).}
\label{fig:intro}
\end{figure}
\begin{figure*}[ht]
\label{table:neighbor_definition}
    \centering
    \begin{subtable}{\textwidth}
        \centering
        \resizebox{\textwidth}{!}{%
        \begin{tabular}{|l|l|l|}
            \hline
            \multicolumn{1}{|c|}{\textbf{Entity}} & \multicolumn{1}{c|}{\textbf{Question}} & \multicolumn{1}{c|}{\textbf{Answer}} \\
            \hline
            J.\,K. Rowling & When was J.\,K. Rowling born? & July 31, 1965 \\
            \cline{2-3}
            & Which book concludes the Harry Potter series written by J.\,K. Rowling? & Harry Potter and the Deathly Hallows \\
            \hline
            Stephen King & When was Stephen King born? & September 21, 1947 \\
            \cline{2-3}
            & Which Stephen King novel features a killer clown named Pennywise? & It \\
            \hline
        \end{tabular}%
        }
        \caption{Examples of Forget Set.}
        \label{tab:forget_set_example}
    \end{subtable}
    
    \vspace{1em} 
    
    \begin{subtable}{\textwidth}
    \centering
    \small 
    \resizebox{\textwidth}{!}{%
    \begin{tabular}{|l|p{0.37\textwidth}|p{0.3\textwidth}|p{0.17\textwidth}|}
        \hline
        \multicolumn{1}{|c|}{\makecell[c]{\textbf{Neighbor Set Type}}} & 
        \multicolumn{1}{c|}{\makecell[c]{\textbf{Entity example}}} & 
        \multicolumn{1}{c|}{\makecell[c]{\textbf{Example Question}}} & 
        \multicolumn{1}{c|}{\makecell[c]{\textbf{Example Answer}}} \\
        \hline
        \makecell[c]{Domain\\Neighbor Set} & 
        \makecell{Mark Twain\\(a writer, of the same profession)} & 
        On which river did Mark Twain work as a steamboat pilot? & Mississippi River \\
        \hline
        \makecell[c]{Entity\\Neighbor Set} & 
        \makecell{Emma Watson\\(the lead actress in Rowling's works)} & 
        What was Emma Watson's first lead role? & Harry Potter \\
        \hline
        \makecell[c]{Syntactically Similar\\Neighbor Set} & 
        \makecell{When was X born?\\(similar syntactic structure as in the forget set) } & 
        When was Nelson Mandela born? & July 18, 1918 \\
        \hline
    \end{tabular}%
    }
    \caption{Examples of Types of Neighbor Sets.}
    \label{tab:neighbor_sets}
\end{subtable}

    \caption{(a) An example forget set consisting of two entities with two QA pairs each; (b) Examples for the three types of neighbor sets: Domain, Entity, and Syntactically Similar.}
    \label{fig:tables_for_unlearning}
\end{figure*}
Large language models (LLMs), trained on vast text corpora, exhibit remarkable capabilities~\citep{dubey2024llama3}. However, their deployment raises concerns about retaining unauthorized content, posing risks in copyright~\citep{karamolegkou2023copyrightviolation}, privacy~\citep{neel2023privacy}. These issues are critical under regulations like GDPR~\citep{voigt2017gdpr}, which mandates post-training data removal and the right to erasure.

To address these challenges, language model unlearning~\citep{yao2023llmunlearningsurvey} has emerged as a promising approach. It aims to achieve two primary objectives. First, the model should effectively forget the information in the forget set, such as private data.  Second, the unlearning process should preserve the model's ability to perform well on tasks unrelated to the forget set, which is represented by the retain set - the remaining subset of the training data that excludes the forget set. Many studies have primarily focused on the first objective, proposing methods to effectively remove the forget set~\citep{sinha2024unstar, eldan2023swhosharrypotter}, or developing metrics to verify whether forgetting has been successful~\citep{lynch2024eight, hu2024joggingthememory}. 
However, unlearning is still rarely used in practice because it is difficult to maintain performance on the retain set.

In this paper, we take a closer look at which areas of the retain set are significantly affected by unlearning through a case study on entity unlearning. Entity unlearning~\citep{maini2024tofu,rwku} aims to remove knowledge about particular entities, typically expressed through QA pairs. 
Since it is not practical to test the whole retain set, previous work has used smaller groups called neighbor sets~\cite{choi2024opt, closerlookat}.
These neighbor sets have similar properties to the data being removed, but they do not include the target data.
They are particularly important as they are expected to experience significant performance degradation during the unlearning process. 
Building on previous work, we conduct an in-depth analysis of these neighbor sets and address two key research questions:
\begin{enumerate}[wide, labelwidth=!, label={\textbf{RQ\arabic*.}}, labelindent=0pt, topsep=2pt, itemsep=-1pt, itemindent=0pt, leftmargin=*, before=\setlength{\listparindent}{-\leftmargin}]
\item How does performance degradation vary across different neighbor sets? (\S\ref{sec:problem})
\item What is the optimal neighbor set for effective regularization?  (\S\ref{sec:solution})
\end{enumerate}

To answer the research questions, we first challenge the conventional approach to neighbor set construction. 
While previous work~\cite{choi2024opt, closerlookat} primarily focused on \textit{Domain Neighbor Sets} containing instances from the same professional domain and \textit{Entity Neighbor Sets}~\cite{rwku, choi2024opt} comprising closely associated entities,
our research reveals that one key factor has been overlooked:  \textit{Syntactic Similarity}. To address this, we introduce the \textit{Syntactically Similar Neighbor Set}, which contains questions sharing similar syntactic structures with the forget set. 
Our experiments show that this set suffers a much larger drop in performance compared to the traditional neighbor sets. (\S\ref{sec:problem}). 
This finding challenges the previous belief~\cite{closerlookat} that entity or domain similarity is the main driver of forgetting patterns. Moreover, the performance degradation is even more pronounced when syntactic similarity overlaps with entity or domain similarity, suggesting a compounding effect. Our paraphrasing experiments and gradient analysis confirm this result by revealing stronger interdependencies within syntactically similar information.

Building on this insight, we evaluate different retain set configurations for regularization during unlearning. 
Despite conventional wisdom~\citep{choi2024opt} suggesting that domain or entity-based retain sets would be most effective, our results demonstrate that training with \textit{Syntactically Similar Neighbor Set} not only best preserves performance on syntactically similar cases but also but also performs as well or better on other parts of the retain set. (\S\ref{sec:solution}). 
This indicates that syntactic similarity, rather than domain or entity relationships, provides a more reliable foundation for maintaining model utility while ensuring effective unlearning.

\section{Preliminaries}
\subsection{Language Model Unlearning}
Let a LLM be parameterized by $\bm{\theta}$ and trained on a dataset $\mathcal{D}$, which consists of a forget set $\mathcal{D}_f$ and a retain set $\mathcal{D}_r = \mathcal{D} \setminus \mathcal{D}_f$. The goal of unlearning is to obtain a new set of parameters $\bm{\theta'}$ that removes knowledge from $\mathcal{D}_f$ while preserving performance on $\mathcal{D}_r$.
\subsection{Entity Unlearning}
Entity unlearning~\citep{maini2024tofu, rwku} aims to remove knowledge associated with specific entities from the LLM. Let $\mathcal{E} = \{e_1, ..., e_n\}$ represent the set of entities to be forgotten, where each entity $e_i$ is represented by a collection of question-answer pairs: $e_i = \{(q_{i,1}, a_{i,1}), ..., (q_{i,m}, a_{i,m})\}$. Thus, the forget set can be expressed as $\mathcal{D}_f = \bigcup_{i=1}^{n} \bigcup_{j=1}^{m} (q_{i,j}, a_{i,j})$.

\subsection{Evaluating Retain Set Preservation}
\label{subsec:EvaluatingRetainSetPreservation}
Since $\mathcal{D}_r$ comprises the entire training set except for $\mathcal{D}_f$, evaluating all of $\mathcal{D}_r$ is impractical. Prior work~\cite{maini2024tofu, rwku} addresses this challenge through two main approaches. First, they assess performance on general knowledge benchmarks such as MMLU~\citep{hendrycks2021measuring} to ensure broad knowledge retention. Second, they evaluate on neighbor sets, which are subsets of $\mathcal{D}_r$ that are expected to be most affected by the unlearning process. These sets are constructed based on the assumption that data points similar to the forget set are more likely to be impacted during unlearning.
Previous work has identified two primary types of neighbor sets:

\noindent
\textbf{Domain Neighbor Set} ($\mathcal{N}_{\text{domain}}$): Instances related to the same professional domain as the forget set~\citep{closerlookat, maini2024tofu}. For example, if $\mathcal{D}_f$ consists of data about J.K. Rowling, $\mathcal{N}_{\text{domain}}$ may include information about other authors such as Ian McEwan.

\noindent
\textbf{Entity Neighbor Set} ($\mathcal{N}_{\text{entity}}$): Instances involving entities closely associated with those in $\mathcal{D}_f$~\citep{rwku, choi2024opt}. For example, if J.K. Rowling is in $\mathcal{D}_f$, then $\mathcal{N}_{\text{entity}}$ may include information about Daniel Radcliffe, the lead actor in the Harry Potter films.

Expanding on the concept of neighbor sets, we propose a new type of neighbor set based on syntactic similarity. 
While existing neighbor sets rely mainly on topical or entity relationships, we observe that performance degradation can also affect instances that share similar syntactic structures. 
We define the \textbf{Syntactically Similar Neighbor Set} ($\mathcal{N}_{\text{syntactically}}$) as a subset of $\mathcal{D}_r$ containing questions with syntactic structures similar to those of $\mathcal{D}_f$. For example, if $\mathcal{D}_f$ contains multiple instances of the form “When was X born?”, $\mathcal{N}_{\text{syntactically}}$ consists of similarly structured questions.

To construct $\mathcal{N}_{\text{syntactically}}$, we use a two-step process that quantifies syntactic similarity between sentences. First, we perform entity masking using GPT-4o~\citep{hurst2024gpt4o} to replace named entities such as person names, dates, and organization names. This allows us to focus on the structural aspects of the sentences while minimizing the influence of specific entities. Let $s_1'$ and $s_2'$ represent the masked versions of sentences $s_1$ and $s_2$, respectively.
Next, we define the Levenshtein similarity based on the Levenshtein distance between the masked sentences. The Levenshtein distance~\cite{LevenshteinDistance} measures the minimum number of edit operations (insertions, deletions, or substitutions) needed to transform one string into another. We normalize this distance into a similarity score using:
\begin{equation}
\label{eq:LevenshteinSimilarity}
\resizebox{0.89\hsize}{0.055\hsize}{$
\text{LevenshteinSimilarity}(s_1, s_2) = 1 - \frac{\text{LevenshteinDistance}(s_1', s_2')}{\max(\text{len}(s_1'), \text{len}(s_2'))}$}
\end{equation}
\begin{algorithm}
\small
\caption{\small Syntactically Similar Neighbor Set Construction}
\begin{algorithmic}[1]
  \Require \parbox[t]{\dimexpr\linewidth-\algorithmicindent}{%
  Set of questions in forget set $\mathcal{D}_f$, $\mathcal{D}_r$, \\
  similarity threshold $\theta_{high}$}
  \Ensure $\mathcal{N}_{syntactically}$
\vspace{0.2pt}
\State Initialize empty set $\mathcal{N}_{syn} \gets \emptyset$
\State Initialize empty clusters $C \gets \emptyset$

\For{each question $q_i, q_j \in \mathcal{D}_f$}
    \State Compute Levenshtein similarity $sim(q_i, q_j)$
    \If{$sim(q_i, q_j) \geq \theta_{high}$}
        \State Group $q_i, q_j$ into same cluster in $C$
    \EndIf
\EndFor

\For{each valid cluster $c \in C$ with size $\geq 3$}
    \State \parbox[t]{\dimexpr\linewidth - \algorithmicindent}{
    Select entities $E$ from retain set not in other neighbor sets
    }
    \vspace{0.2pt}
    \State \parbox[t]{\dimexpr\linewidth - \algorithmicindent}{
      Generate QA pairs for $E$ with similar syntactic \\
      structure
    }
    \vspace{0.2pt}
    \State Verify generated pairs via model probing
    \State Add verified pairs to $\mathcal{N}_{syn}$
\EndFor

\State \Return $\mathcal{N}_{syn}$
\end{algorithmic}
\label{algorithm:constructingDataset}
\end{algorithm}

\section{Dataset Construction}
We consider two scenarios for entity unlearning: the fictitious author scenario (TOFU) and a real-world scenario involving actual individuals.  This section details the construction of the forget set and the various neighbor sets for each scenario.

\subsection{Target Entity Selection}

For the real-world scenario, we first select 10 prominent figures across professions: actors, singers, politicians, and business leaders, etc.  These individuals are chosen based on their public visibility and the availability of information about them~\citep{rwku, choi2024opt}. In the TOFU scenario, we follow the method outlined in~\citet{maini2024tofu}, employing a 1\% forget ratio to determine the number of fictitious authors to be included in the forget set.

\subsection{Neighboring Entity Selection}
\label{sec:neighborentityselection}
The selection process for each type of neighbor set varies depending on the specific criteria for each.

\paragraph{Domain Neighbor Set.} For the real-world scenario, domain neighbor entities are constructed by selecting individuals within the same professional domain as the target entities following~\citet{closerlookat, liu2024revisitingharrypotter}. In the TOFU scenario, the domain neighbors provided in~\citet{maini2024tofu} are used.
\paragraph{Entity Neighbor Set.}  For the real-world scenario, entity neighbor entities are selected based on the following criteria adapted from~\citet{choi2024opt, rwku}: 1) a bidirectional relationship exists between the target entity and the potential neighbor, meaning both entities link to each other via hyperlinks on their respective Wikipedia pages and are mentioned at least once on those pages; and 2) the neighboring pages all represent people.  These criteria aim to identify entities closely associated with the target entities, reflecting real-world relationships and connections. For the TOFU scenario, given its fictitious nature, and the absence of a defined entity neighbor concept in~\citet{maini2024tofu}, entity neighbors are not applicable.

\paragraph{Syntactically Similar Neighbor Set.} Unlike the other neighbor sets, which are based on entities, the syntactically similar neighbor set is constructed using questions in $\mathcal{D}_f$. This set consists of questions in the retain set that share a similar syntactic structure with those in the $\mathcal{D}_f$. To construct this set, we first compute the pairwise Levenshtein similarity, as defined in equation~\ref{eq:LevenshteinSimilarity}, between all questions in $\mathcal{D}_f$. Then, we group questions ensuring that each question within a cluster is syntactically similar to the others in that cluster.

\subsection{Generating QA Pairs}
Based on the selected entities, we generate QA pairs that capture key information about each entity.
\paragraph{Real-world Scenario.}
We utilize Wikipedia as a knowledge source following~\citet{rwku}.

For the forget set, domain neighbor set, and entity neighbor set, we employ GPT-4o to generate QA pairs for each entity.  We first gather relevant passages from Wikipedia pages corresponding to each target entity. These passages serve as the context for prompting GPT-4o to generate QA pairs related to the targets. Second, we further filter the QA pairs by prompting GPT-4o with the questions alone—without any passages—and retaining only those for which it produces the correct answer. 

To validate the model's knowledge and the quality of the generated pairs, we use these QA pairs to probe the evaluated model. We retain only those QA pairs for which the model successfully recalls the correct answer. This validation ensures both the consistency of the QA pairs and confirms the model's existing knowledge. 

For constructing the syntactically similar neighbor set, we first identify entities from the retain set that are not included in any of the other neighbor sets (forget, domain, or entity). Using the syntactic clusters identified in Section~\ref{sec:neighborentityselection}, we generate QA pairs that align with the syntactic structures of these clusters.

Specifically, we adopt the masking approach used in Section~\ref{subsec:EvaluatingRetainSetPreservation} when computing Levenshtein similarity. We first mask entity within the clustered questions and then generate new QA pairs by filling these masked structures with entities from the identified retain set. This ensures that the generated questions maintain syntactic similarity to the existing clusters while introducing new entities. We follow the same verification process (model probing and manual verification) as for the other neighbor sets to ensure the dataset's validity. The detailed procedure for constructing the syntactically similar neighbor set is outlined in Algorithm~\ref{algorithm:constructingDataset}.

\paragraph{TOFU.} For the TOFU, the forget set and domain neighbor entities are defined by the benchmark itself~\citep{maini2024tofu}. To identify the syntactically similar neighbor set, we compare the provided neighbor sets against the forget set using the same syntactic similarity clustering method described above. Critically, we ensure that there is no overlap with the domain neighbor set. This approach ensures that the syntactically similar neighbor set reflects the structural patterns present in the forget set while maintaining distinctness from other neighbor sets. 

Further details and dataset statistics are provided in the appendix~\ref{appendix:dataset_construction}.
\section{Experimental Setup}
\subsection{Evaluation Metrics}
We evaluate the unlearned model using several metrics to assess its performance from various perspectives~\citep{closerlookat, maini2024tofu}. Specifically, we employ \emph{ROUGE} to measure word-level similarity, \emph{BERT Cosine Similarity} to assess semantic consistency between outputs before and after unlearning, \emph{Probability} to evaluate the model's confidence to predict the ground truth answer, and \emph{Entailment Score} to assess factual correctness relative to the ground truth.

\noindent
Since all metrics range from zero to one, we aggregate them using the arithmetic mean. Applying this to the retain set defines \textbf{Model Utility (MU)}, while applying it to the forget set defines \textbf{Forget Efficacy (FE)}.

\noindent
To quantify the impact of unlearning on neighbor sets, we define the \textbf{Relative Utility Drop (RUD)} as:
\begin{equation}
\resizebox{0.6\hsize}{!}{$
\textstyle \text{RUD} = \frac{MU_{\text{after}} - MU_{\text{before}}}{MU_{\text{before}}} \times 100.$}
\end{equation}
Since unlearning typically reduces MU, RUD is usually negative, indicating the degree of performance drop. This metric shows which neighbor set suffers the most performance decline after unlearning. Further details on metric computation are provided in Appendix~\ref{appendix:evaluationMetrics}.
\subsection{Unlearning Methods}
We explore various unlearning strategies, each of which aims to erase knowledge of target entities in distinct ways. A comprehensive explanation of these methods is provided in Appendix~\ref{appendix:overviewUnlearningMethods}.
\begin{itemize}[leftmargin=6pt]
    \item \textbf{GA}~\cite{jang2023knowledgeunlearning}: Utilizes gradient ascent on the forget set to counteract learned knowledge.
    \item \textbf{DPO}~\cite{rafailov2023dpo}: Treats unlearning as a preference optimization problem by applying the standard DPO loss. It uses answers in the forget set as negative samples and rejection templates (e.g., ``I don't know'') as positive samples to guide the model's response.
    \item \textbf{NPO}~\cite{zhang2024npo}: A variant of DPO that removes positive samples from the optimization process. It retains only negative examples from the forget set, encouraging the model to suppress forgotten information without explicit reinforcement of alternative responses.
    \item \textbf{IDK}~\cite{maini2024tofu}: Fine-tunes the model to default to ``I don't know'' responses when queried about the forget set.
\end{itemize}

\subsection{Implementation Details}
\begin{table}[h]
    \centering
    \begin{tabular}{lcccc}
        \hline
        & \textbf{GA} & \textbf{NPO} & \textbf{IDK} & \textbf{DPO} \\
        \hline
        \textbf{Real-world} & 0.734 & 0.745 & 0.657 & 0.721 \\
        \textbf{TOFU} & 0.676 & 0.710 & 0.685 & 0.686 \\
        \hline
    \end{tabular}
    \caption{Forget efficacy of each method across different scenarios.}
    \label{tab:forget-efficacy}
\end{table}

For the TOFU benchmark~\citep{maini2024tofu}, we utilize fine-tuned Llama-2-7b-chat~\citep{touvron2023llama2}, which has been trained on the constructed dataset to ensure it precisely answers questions in TOFU. For the real-world scenario benchmark, we employ Llama-3-8B-Instruct~\citep{dubey2024llama3}.
\noindent
To enable a fair comparison of different unlearning methods at similar levels of forgetting, we adjust the hyperparameters to keep Forget Efficacy between 0.65 and 0.75. Further details are provided in Appendix~\ref{appendix:detailedResultsPerMethods}.
\begin{figure*}[t]
    \centering
    \begin{subfigure}{0.49\linewidth}
        \includegraphics[width=\linewidth]{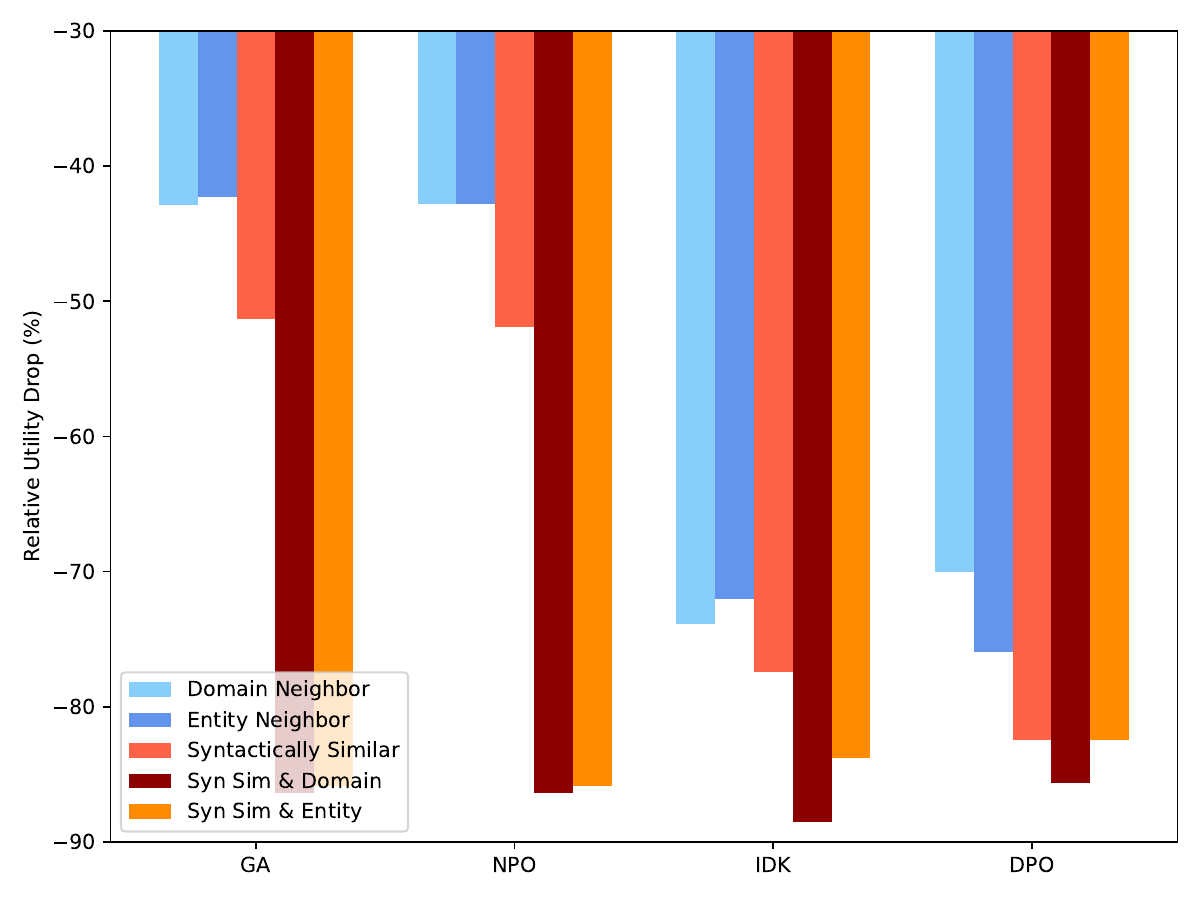}
        \caption{Real-world Scenario}
        \label{fig:real-world_main}
    \end{subfigure}
    \hfill
    \begin{subfigure}{0.49\linewidth}
    \label{fig:tofu_main}
        \includegraphics[width=\linewidth]{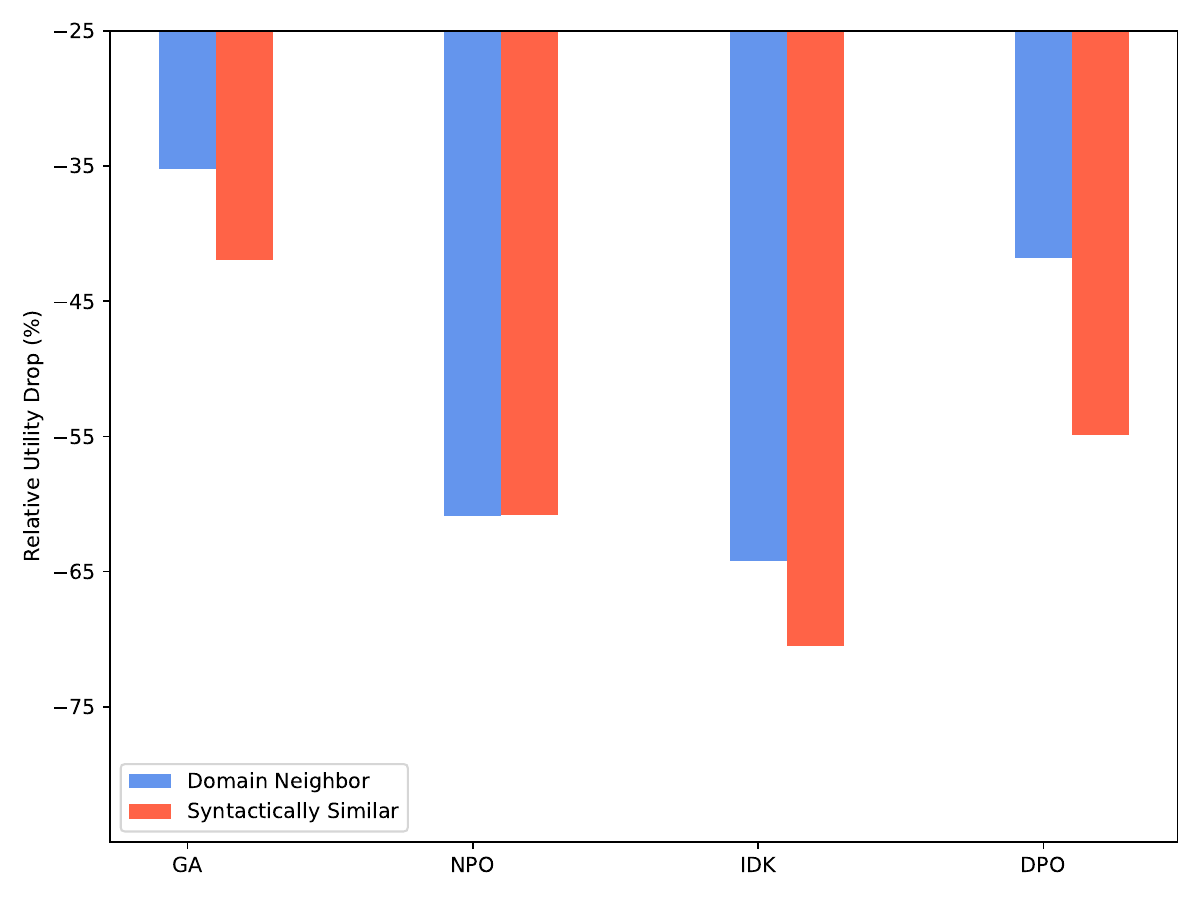}
        \caption{TOFU}
        \label{fig:tofu_main}
    \end{subfigure}
    \caption{Relative Utility Drop (\%) for different neighbor sets across real-world scenario (left) and TOFU (right). Each method (GA, NPO, IDK, DPO) is evaluated based on its model utility before and after unlearning, with lower bars indicating greater utility loss. Model utility values before and after unlearning are provided in Appendix~\ref{appendix:detailedResultsPerMethods}}
    \label{fig:experiment1}
\end{figure*}

\section{How does Performance Degradation Vary across Different Neighbor Sets?}
\label{sec:problem}
This section investigates how performance degradation after unlearning varies across different neighbor sets.
First, we examine which neighbor sets experience the most significant performance degradation. (Section~\ref{sec:5_1}) If similar syntactic structures sets are the most vulnerable to forgetting, we further examine whether domain differences within these structures lead to varying effects. (Section~\ref{sec:5_2}) We then examine the robustness of these forgetting patterns when questions are paraphrased. (Section~\ref{sec:5_3}) Finally, we analyze gradient updates during unlearning to understand the underlying mechanisms driving the observed patterns. (Section~\ref{sec:5_4})

\subsection{Analyzing Performance Drops Across Neighbor Sets}
\label{sec:5_1}
\vspace{2pt}
\textbf{Syntactically Similar Neighbor Set Experiences Higher Forgetting.} Across both real-world scenario and  TOFU evaluations (Figure~\ref{fig:real-world_main} and Figure~\ref{fig:tofu_main}), $\mathcal{N}_{\text{syntactically}}$ consistently demonstrates a higher utility drop compared to both $\mathcal{N}_{\text{domain}}$ and $\mathcal{N}_{\text{entity}}$. The greater utility drop suggests that syntactic similarity plays a crucial role in the forgetting phenomenon.  When the model is unlearning specific data, it appears to struggle more with retaining information that shares similar sentence structures, regardless of the specific domain or entities involved.

\noindent \textbf{No Significant Difference among Existing Neighbor Sets.} In the real-world scenario, a notable observation is the lack of significant performance differences between $\mathcal{N}_{\text{domain}}$ and $\mathcal{N}_{\text{entity}}$. As depicted in Figure~\ref{fig:real-world_main}, both sets exhibit similar RUD across all methods. Our results show that, despite different ways of defining neighbor sets in previous studies~\citep{choi2024opt, closerlookat}, the impact caused by unlearning is similar across them.

\noindent \textbf{Overlapping Sets Lead to Even Greater Forgetting.}  
In the real-world scenario, subsets that overlap syntactic similarity with domain or entity similarity (\textit{Syn Sim \& Domain}, \textit{Syn Sim \& Entity}) experience the most severe utility drop (Figure~\ref{fig:real-world_main}). This highlights that overlapping neighbor characteristics intensify forgetting effects during unlearning.

\subsection{Exploring Domain Effects on Forgetting in Syntactically Similar Cases}
\label{sec:5_2}
To examine the domain-specific effects of unlearning in syntactically similar cases, we conduct experiments in real-world scenario across five distinct categories. This analysis builds on our previous findings that syntactically similar neighbor sets exhibit more pronounced forgetting than those based on domain or entity similarity.

While overlapping characteristics intensify forgetting, this raises the question of which similarity type is the primary driver. Prior studies~\cite{rwku,maini2024tofu} have operated on the assumption that entity or domain similarity is the most critical factor, meaning sets with high internal similarity would be most vulnerable. Following this logic, the \textit{Human} category, containing closely related entities, should exhibit the highest degree of forgetting.

\begin{figure}[t]
    \centering
    \includegraphics[width=1\linewidth]{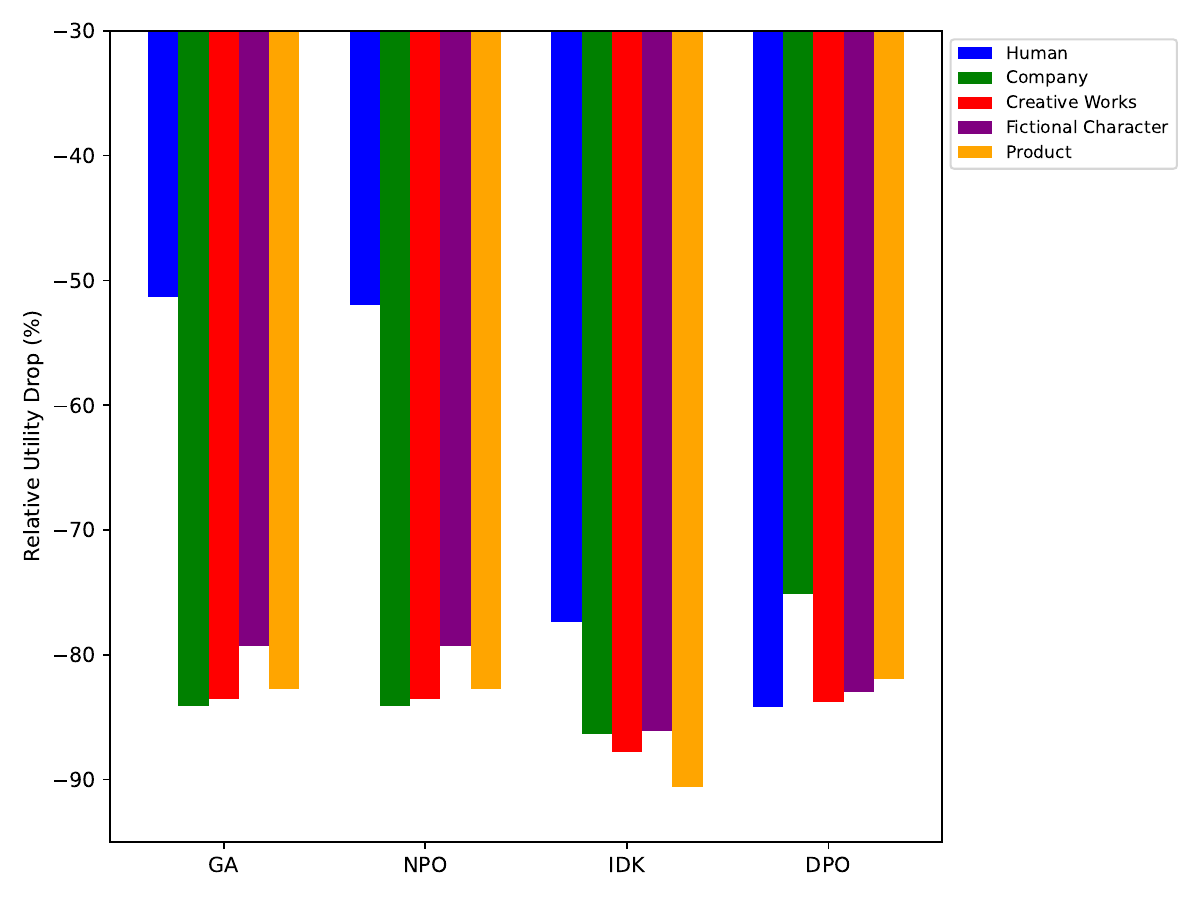}
    \caption{Relative Utility Drop across different entity categories (Human, Company, Creative Works, Fictional Character, and Product) for various unlearning methods.}
    \label{fig:variousfig}
\end{figure}

However, as shown in Figure~\ref{fig:variousfig}, the results trend in the opposite direction—non-human categories consistently exhibit substantially higher forgetting rates across most methods. This directly challenges the conventional assumption that entity or domain similarity is the most reliable predictor of performance degradation. Instead, it suggests that these factors are secondary to a more influential driver, reinforcing our central claim about the overriding importance of syntactic structure.
\begin{figure}[t]
    \centering
    \includegraphics[width=1\linewidth]{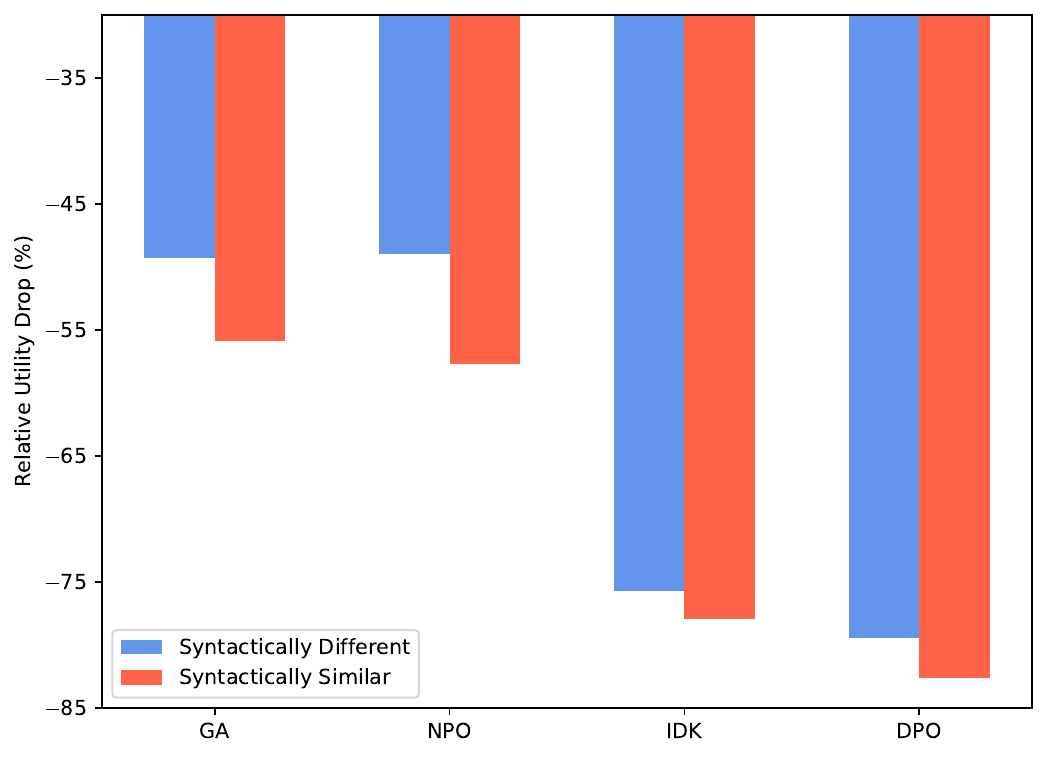}
    \caption{Relative Utility Drop for syntactically similar and different neighbor sets across different unlearning methods, measured over three paraphrases per question. A larger drop indicates higher semantic forgetting.}
    \label{fig:paraphrasefig}
\end{figure}
\subsection{Robustness of Forgetting Patterns in Paraphrased Scenarios}
\label{sec:5_3}
Our previous experiments reveal that syntactically similar neighbor sets experience higher levels of forgetting compared to other neighbor sets. To validate the robustness of this finding, we investigate whether this performance gap persists even when questions are paraphrased. 

Specifically, we generate paraphrased versions for each question for syntactically similar and different neighbor sets using GPT-4o Then, we filter out cases where the pre-unlearning model fails to provide correct answers, ensuring that each question has three valid paraphrases. We then measure the RUD for these paraphrased questions using the post-unlearning model and compare the forgetting rates across the two groups.

Figure~\ref{fig:paraphrasefig} shows that even after paraphrasing, syntactically similar neighbors exhibit greater utility drops than dissimilar neighbors. This suggests that the model's increased forgetting isn't solely due to shared syntax, but also reflects a sensitivity to underlying semantic relationships. The consistent performance gap after paraphrasing reinforces the role of syntactic similarity in forgetting, highlighting its influence beyond surface-level wording.


\subsection{Gradient Analysis}
\label{sec:5_4}
To further investigate the underlying mechanisms behind the forgetting patterns observed in syntactically similar and dissimilar neighbor sets, we analyze the gradient updates during the unlearning process. Our primary goal is to understand how the model's gradient norms evolve when encountering different types of neighbors, particularly whether syntactically similar instances influence each other more strongly than dissimilar ones.

In our experimental setup, we perform gradient ascent on a syntactically similar set and track the changes in gradient norms as the model encounters other syntactically similar or syntactically different instances. Specifically, we measure the Frobenius norm of the model’s weight gradients at each unlearning step, comparing how the gradients behave when interacting with different types of data points.

\begin{figure}[t]
    \centering
    \includegraphics[width=1\linewidth]{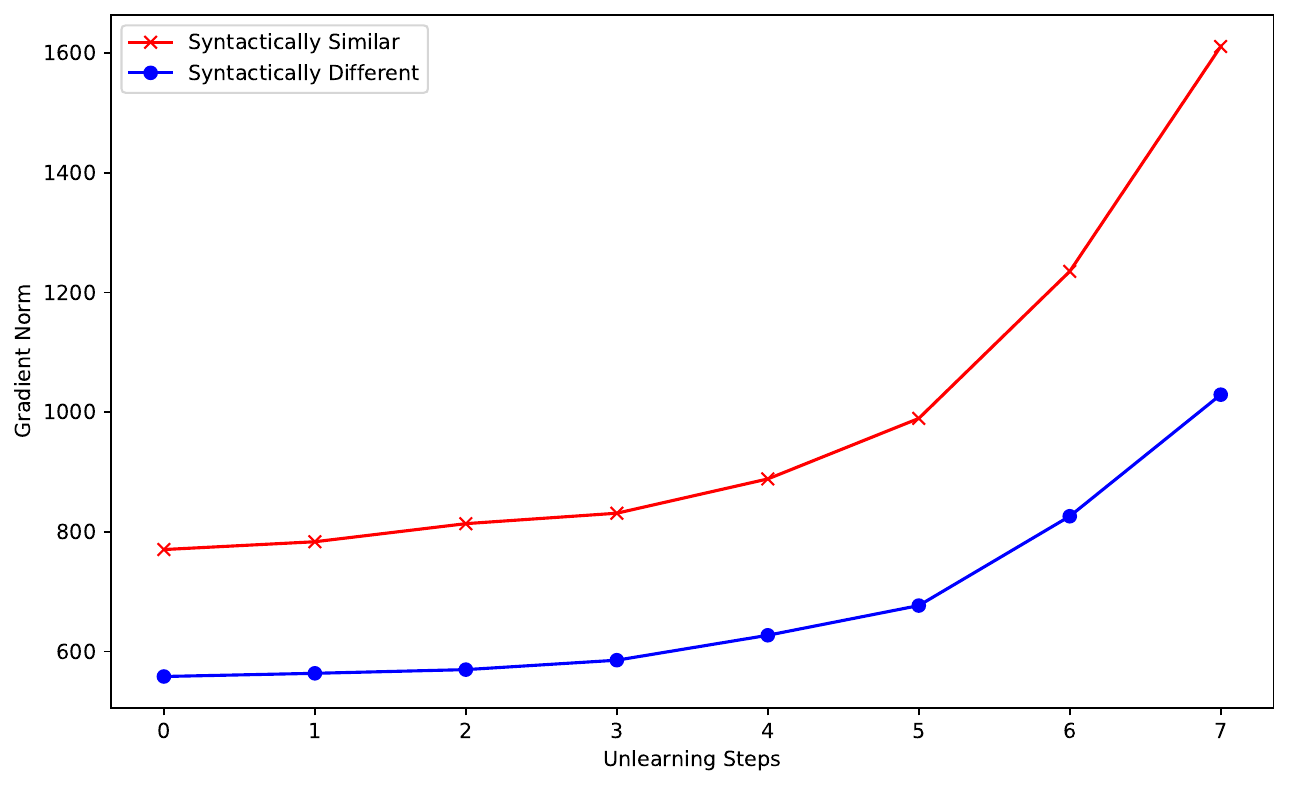}
    \caption{Frobenius norm of model weight gradients across unlearning steps. The gradient norms for syntactically similar instances (red) increase more steeply than those for syntactically different instances (blue).}
    \label{fig:gradientanalysis}
\end{figure}
As shown in Figure~\ref{fig:gradientanalysis}, the gradient norms of syntactically similar instances exhibit a steeper increase over unlearning steps compared to syntactically different instances. Notably, the initial gap between their gradient norms at the first checkpoint widens progressively as unlearning proceeds. This suggests that forgetting syntactically similar knowledge amplifies gradient updates in a way that reinforces the distinction between similar and dissimilar instances.
\begin{figure*}[t]
    \centering
    \includegraphics[width=0.9\linewidth]{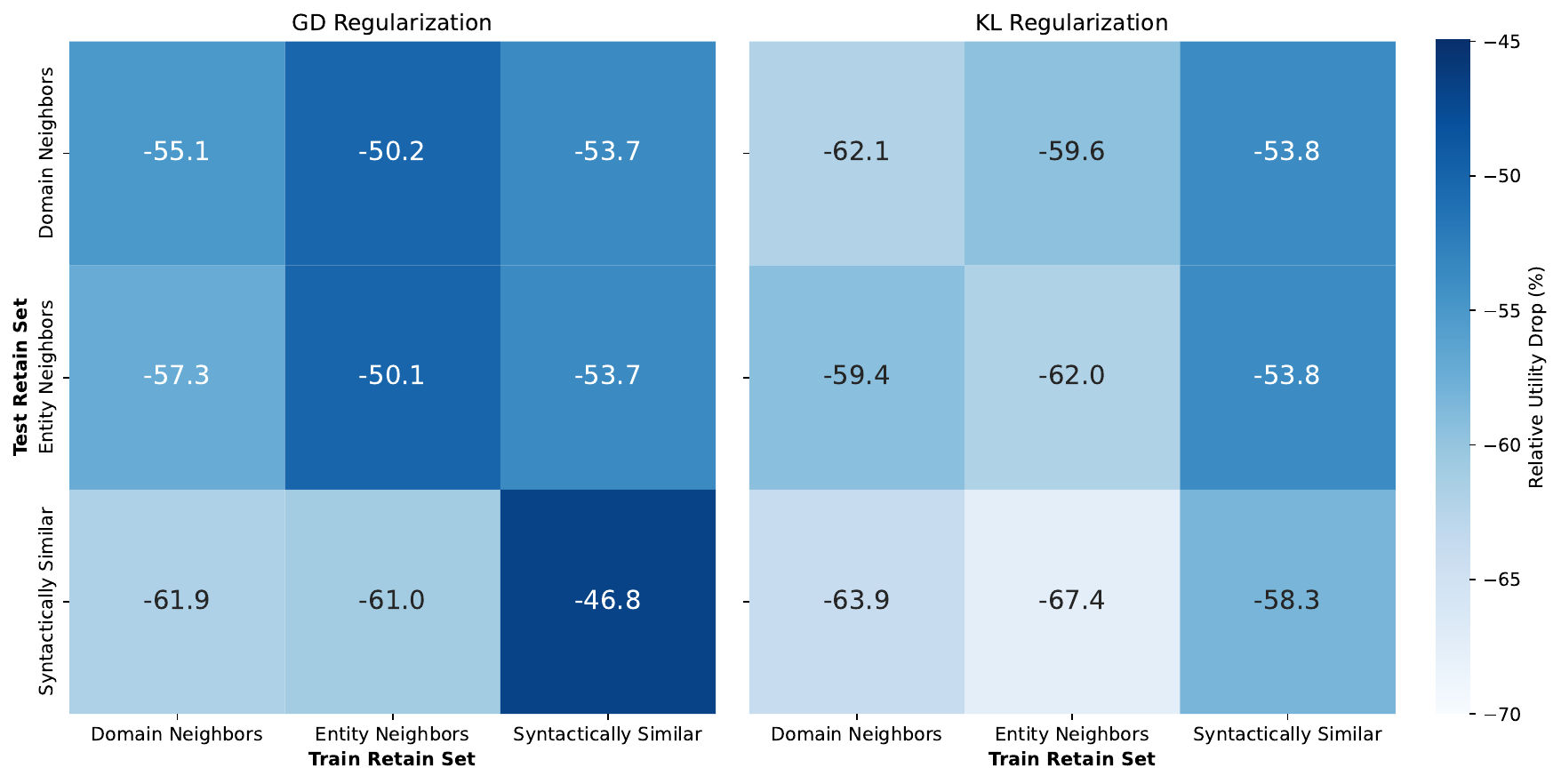}
    \caption{Relative utility drop (\%) averaged across all unlearning methods (GA, DPO, NPO, and IDK) under different retain set configurations using GD (left) and KL (right) regularization. The x-axis represents the type of train retain set, while the y-axis represents the type of test retain set. A higher value (darker color) indicates better utility retention. Detailed relative utility drop results for each individual unlearning method can be found in Appendix~\ref{appendix:detailedResultsPerMethods}.}
    \label{fig:regularizationheapmap}
\end{figure*}
\section{What is the Optimal Neighbor Set for Effective Regularization?}
\label{sec:solution}
To preserve model utility during unlearning, regularization losses on a subset of the retain set are commonly employed during the unlearning process~\cite{closerlookat,maini2024tofu}. Based on the findings of the previous section, we aim to identify the optimal configuration of the retain set used for regularization, to optimize model utility while ensuring successful forgetting, specifically from a data perspective.

\paragraph{Regularization loss.} It encourages the unlearned model parameters $\bm \theta$ to preserves model utility. A typical unlearning objective function, computed on a subset of $\mathcal{D}_{\text{R}}$, is formulated as follows:
\begin{equation}
    \resizebox{0.71\hsize}{!}{$
    \underset{\bm \theta}{\min} \mathcal{L}(\bm \theta) = \underset{\bm \theta}{\min} - \mathcal{L}_f(\bm \theta) +  \mathcal{L}_{\text{R}}(\theta;\mathcal{D}_{\text{R}}).$}
    \label{eq:previous_objective}
\end{equation}
Our analysis considers two primary regularization approaches: Gradient Descent (GD) and Kullback-Leibler Divergence (KL). A comprehensive explanation of these methods is provided in Appendix~\ref{appendix:overviewUnlearningMethods}.
\paragraph{Setup.} To determine the optimal train retain set configuration, we conduct comprehensive experiments examining nine different combinations of train and test retain sets, using $\mathcal{N}_{\text{domain}}$, $\mathcal{N}_{\text{entity}}$, and $\mathcal{N}_{\text{syntactically}}$ for both training and evaluation. For each train retain set, we apply different unlearning methods (GA, DPO, NPO, and IDK) with regularization loss and report the average RUD across test retain sets.

\paragraph{Results.} We visualize the results separately for GD and KL regularization in Figure~\ref{fig:regularizationheapmap}. The results reveal two key findings:

\noindent
\textbf{1) Training with $\mathcal{N}_{\text{syntactically}}$ effectively preserves performance on $\mathcal{N}_{\text{syntactically}}$.} In both GD and KL regularization heatmaps, the bottom row (Test Retain Set: Syntactically Similar) shows that training with $\mathcal{N}_{\text{syntactically}}$ preserves utility best, with average differences of 14.7\% point and 7.35\% point compared to other training sets, respectively.

\noindent
\textbf{2) Training with $\mathcal{N}_{\text{syntactically}}$ contributes to robust performance across various neighbor sets.} Beyond preserving performance on syntactically similar data, training with $\mathcal{N}_{\text{syntactically}}$ also yields competitive results when evaluated on $\mathcal{N}_{\text{entity}}$ and $\mathcal{N}_{\text{domain}}$. In many cases, it surpasses or closely matches the performance achieved by training with other neighbor sets. These findings emphasize the role of syntactically similar examples in reducing utility loss while unlearning.
\section{Related Work}
LLM unlearning~\citep{jang2023knowledgeunlearning, yao2023llmunlearningsurvey, lynch2024eight} has gained significant attention as a method to enhance privacy. Various approaches~\citep{sinha2024unstar, zhang2024npo} have been proposed to ensure that models effectively erase specific information while maintaining overall performance. A key challenge in unlearning is assessing whether knowledge unrelated to the forget set is inadvertently affected. To evaluate this, researchers commonly examine general knowledge~\citep{hendrycks2021measuring, cobbe2021training} as well as a designated subset of the retain set that shares a similar distribution with the forget set but excludes the targeted information. These subsets, often referred to as neighbor sets~\citep{closerlookat}, help determine the extent of unintended degradation in model performance.

In hazardous knowledge unlearning, prior work has leveraged domain-relevant general knowledge as a benchmark. For instance,~\citet{li2024wmdp} employs general biology knowledge to assess the impact of bioweapon-related unlearning and general computer security knowledge to evaluate the removal of information related to Attacking Critical Infrastructure. For entity unlearning~\citep{maini2024tofu, rwku}, previous studies have used entities from similar professions or those closely linked to the target entity as neighbor sets. While these approaches provide an initial framework, they lack a systematic investigation of which aspects of the retain set suffer the most from unlearning. Our study addresses this gap by systematically investigating the impact of unlearning on different types of neighbor sets more clearly and identifying which knowledge components experience the highest degree of forgetting.
\section{Conclusion}
In this paper, we examine unlearning's impact on retain sets and highlight the Syntactically Similar Neighbor Set as key to forgetting patterns. Our results show syntactic similarity, not domain or entity ties, drives retained knowledge degradation. Experiments confirm that syntactically similar neighbors face the highest utility drop, challenging prior assumptions. We also find that using such data for regularization improves performance retention. These findings refine unlearning strategies and emphasize the role of syntactic structure in minimizing unintended knowledge loss.

\section*{Limitations}
Our study focuses on entity unlearning, leaving hazardous knowledge and copyrighted content unlearning unexplored. These cases may require different evaluation strategies.

Additionally, our experiments use mid-sized models (LLaMA-2-7B-Chat, LLaMA-3-8B-Instruct). Larger models, with their computational demands and structural differences, may respond differently. Future research should assess their applicability to such models.

\section*{Acknowledgement}
This work was supported by the Institute of Information \& Communications Technology Planning \& Evaluation (IITP) grant funded by the Korea government (MSIT) [RS-2021-II211341, Artificial Intelligent Graduate School Program (Chung-Ang University)].

\bibliography{acl}
\appendix
\newpage
\section{Evaluation Metrics Details}
\label{appendix:evaluationMetrics}This section provides details on the metrics used to assess the effectiveness of unlearning. These metrics capture different aspects of model performance, including lexical similarity, semantic consistency, confidence in predictions, and factual correctness.

\noindent
\textbf{ROUGE} measures how closely the model's output aligns with the ground truth at the word level. Specifically, we use ROUGE-L recall~\citep{lin2004rouge}, which considers the longest common subsequence between the model’s generated output $g(x;\theta_u)$ and the correct answer $y$. This metric is useful for evaluating whether the model retains relevant content after unlearning.

\noindent
\textbf{Probability} quantifies the likelihood that the model correctly predicts the ground truth answer. Following~\citet{maini2024tofu}, we compute the normalized conditional probability of the ground truth, defined as $P(y|x) = \frac{1}{T} \sum_{t=1}^{T} p(y_t|x \circ y_{<t}; \theta_u)$. A lower probability after unlearning indicates reduced model confidence in generating the forgotten content.

\noindent
\textbf{Cosine Similarity} assesses the semantic consistency of model outputs before and after unlearning. Inspired by semantic textual similarity tasks~\citep{cer-etal-2017-semeval}, we embed the outputs using Sentence-BERT~\citep{reimers-2019-sentence-bert} and compute their cosine similarity. We set a lower bound of 0, defining the metric as $\max(\text{Cos}(g(x;\theta), g(x;\theta_u)), 0)$. Lower similarity scores indicate greater divergence in output, often due to additional or altered information introduced post-unlearning.

\noindent
\textbf{Entailment Score} evaluates the factual correctness of generated responses relative to the ground truth. This metric is based on Natural Language Inference (NLI), where a pre-trained NLI model~\citep{sileo2023tasksource} determines whether the model's output logically follows from the reference answer~\citep{liu2024learning}. The final score represents the proportion of outputs classified as “entailment.” Higher values indicate better factual alignment, particularly for retained knowledge, while lower scores suggest effective forgetting of targeted information.

These metrics collectively provide a comprehensive evaluation of the unlearning process by measuring its impact on both forgotten and retained knowledge.
\section{Overview of Unlearning Methods}
\label{appendix:overviewUnlearningMethods}
This section provides a detailed explanation of the unlearning methods discussed in the main text, describing their underlying principles and mathematical formulations.

\subsection{Gradient Ascent (GA)}
Gradient Ascent (GA) directly modifies the model’s behavior by applying optimization in the reverse direction of standard training. The objective function for GA is defined as:
\begin{equation}\label{eq:GA}
\resizebox{0.81\hsize}{!}{$
\mathcal{L}_{\text{GA}}(\mathcal{D}_{\text{F}};\theta) 
= - \mathbb{E}_{(x,y) \sim \mathcal{D}_{\text{F}}} \left[ -\log p(y|x; \theta) \right]. 
$}
\end{equation}
\subsection{Negative Preference Optimization (NPO)}
Negative Preference Optimization (NPO) treats unlearning as a preference optimization problem by discouraging responses associated with the forget set. It adapts Direct Preference Optimization (DPO) by treating answers in the forget set as undesirable and excluding positive terms from the DPO loss. The loss function for NPO is given by:
\begin{equation}\label{eq:NPO}
\resizebox{0.81\hsize}{!}{$
\mathcal{L}_{\text{NPO}}(\mathcal{D}_{\text{F}};\theta) 
= -\frac{2}{\beta} \mathbb{E}_{(x,y)\sim \mathcal{D}_{\text{R}}} \left[ \log \sigma \left(-\beta \log \frac{p(y|x;\theta)}{p(y|x;\theta_{\text{ref}})} \right) \right],
$}
\end{equation}
where $\beta$ is a hyperparameter, and $\theta_{\text{ref}}$ represents the reference model, typically the initial model before unlearning. NPO dynamically adjusts gradient weights, making it an adaptive form of GA.

\subsection{Direct Preference Optimization (DPO)}
Direct Preference Optimization (DPO) formalizes unlearning as a preference ranking problem by contrasting the probabilities of retaining and forgetting knowledge. In this approach, responses from the forget set are treated as negative examples, while predefined rejection responses are treated as positive.

\subsection{IDK Fine-tuning (IDK)}
IDK Fine-tuning reframes unlearning as an instruction-tuning task by relabeling forget set queries with predefined rejection templates. This ensures that the model responds with a standardized “I don’t know” response instead of recalling forgotten information. The objective function is:
\begin{equation}\label{eq:IDK}
\resizebox{0.81\hsize}{!}{$
\mathcal{L}_{\text{IDK}}(\mathcal{D}_{\text{F}},\mathcal{D}_{\text{IDK}};\theta) 
= \mathbb{E}_{x \sim \mathcal{D}_{\text{F}}, y \sim \mathcal{D}_{\text{IDK}}}  \left[ -\log p(y|x; \theta) \right].$}
\end{equation}
where $\mathcal{D}_{\text{IDK}}$ contains multiple rejection templates. By fine-tuning on these templates, the model systematically replaces knowledge recall with a controlled rejection response.
\subsection{Regularization Loss}
While the aforementioned losses focus solely on unlearning, a robust method must also preserve the model’s utility. To achieve this, a regularization loss is applied to the retain set, ensuring that useful knowledge remains intact.

\textbf{Gradient Descent (GD)} directly applies the standard prediction loss to the retain set, reinforcing learned knowledge:
\begin{equation}\label{eq:GD}
\resizebox{0.81\hsize}{!}{$
\mathcal{L}_{\text{GD}}(\mathcal{D}_{\text{R}};\theta) = \mathbb{E}_{{(x,y) \sim \mathcal{D}_{\text{R}}}} \left[ -\log p(y|x; \theta) \right].$}
\end{equation}

\textbf{Kullback-Leibler Divergence (KL)} maintains consistency between the unlearned and reference model predictions by minimizing KL divergence on the retain set:
\begin{equation}\label{eq:KL}
\resizebox{0.86\hsize}{!}{$
\mathcal{L}_{\text{KL}}(\mathcal{D}_{\text{R}};\theta) = \mathbb{E}_{(x,y) \sim \mathcal{D}_{\text{R}}} \left[\text{KL}(p(y|x; \theta) \Vert p(y|x; \theta_{\text{ref}}))\right].$}
\end{equation}

By combining different unlearning objectives with regularization losses, we obtain seven baseline methods: GA+GD, GA+KL, NPO+GD, NPO+KL, DPO+GD, DPO+KL, IDK+GD, and IDK+KL.

\section{Further Implementation Details}
All experiments are conducted on two NVIDIA RTX 6000 Ada GPUs. We utilize DeepSpeed with ZeRO3 to reduce memory costs. The AdamW optimizer is employed with a weight decay of 0.01, and all experiments use an effective batch size of 32.
To ensure a fair comparison across different unlearning methods, we adjust training epochs and the learning rate to maintain a Forget Efficacy within the range of 0.65 to 0.75. This range is selected to establish a common baseline for model utility across methods, ensuring that comparisons are not skewed by differences in the extent of forgetting. 
\label{appendix:implementationDetails}
\begin{table}[h]
    \centering
    \begin{tabular}{lcc}
        \hline
        & \textbf{lr} & \textbf{epochs} \\
        \hline
        GA  & 5.00E-06 & 3 \\
        NPO & 3.00E-05 & 3 \\
        IDK & 3.00E-06 & 2 \\
        DPO & 8.00E-06 & 4 \\
        \hline
    \end{tabular}
    \caption{Hyperparameters of real world scenarios experiments}
    \label{tab:hyperparams-realworld}
\end{table}
\begin{table}[h]
    \centering
    \begin{tabular}{lcc}
        \hline
        & \textbf{lr} & \textbf{epochs} \\
        \hline
        GA  & 2.00E-05 & 4 \\
        NPO & 4.00E-05 & 5 \\
        IDK & 2.00E-05 & 2 \\
        DPO & 4.00E-05 & 2 \\
        \hline
    \end{tabular}
    \caption{Hyperparameters of TOFU experiments}
    \label{tab:hyperparams-tofu}
\end{table}

\section{Detailed Explanation of Syntactically Similar Neighbor Set Construction}
\label{appendix:dataset_construction}

\paragraph{Definition of Syntactic Similarity.} We define syntactic similarity based on the Levenshtein similarity score. Specifically, we consider two questions to be syntactically similar if their Levenshtein similarity is at least 0.75. Conversely, if the similarity is 0.4 or lower, they are deemed syntactically different. These thresholds ensure a clear distinction between syntactically similar and different questions while allowing for slight variations in wording.

\paragraph{Ensuring Syntactic Distinctness in Other Neighbor Sets.} The syntactically similar neighbor set is the only set where elements share syntactic structures with the forget set. To ensure differentiation, all other neighbor sets (i.e., domain neighbor and entity neighbor sets) consist of questions classified as syntactically different (Levenshtein similarity $\leq 0.4$) from those in the forget set. This ensures that these sets are semantically related but do not overlap structurally with the forget set.

\paragraph{Clustering Criteria.} 
Each syntactic cluster is formed such that all elements within it are syntactically similar (Levenshtein similarity $\geq 0.75$). To ensure meaningful groupings, we define a cluster as valid only if it contains at least three elements. This criteria ensure that syntactically similar neighbor sets are well-defined and systematically constructed across both scenarios while maintaining clear distinctions from other neighbor sets.
\begin{table}[h]
    \centering
    \begin{tabular}{lcc}
        \hline
        & \textbf{TOFU} & \textbf{real-world scenario} \\
        \hline
        Forget      & 40   & 150 \\
        \hline
    \end{tabular}
    \caption{Data statistics for different forget sets.}
    \label{tab:data-statistics-forget}
\end{table}

\begin{table}[h]
    \centering
    \begin{tabular}{lcc}
        \hline
        & \textbf{TOFU} & \textbf{real-world scenario} \\
        \hline
        Entity      & 0   & 182 \\
        Domain      & 34  & 150 \\
        SynSimilar  & 34  & 212 \\
        \hline
    \end{tabular}
    \caption{Data statistics for different neighbor sets.}
    \label{tab:data-statistics-neighbor}
\end{table}

\section{Detailed Prompts}
\onecolumn

\begin{figure}[ht]
\begin{tcolorbox}[width=0.9\textwidth, colback=white, colframe=black, boxsep=5pt, fontupper=\small]
Instruction:

Replace specific parts of the text with \verb|{X}| to anonymize information. The specific parts include:

- Person's name
- Date
- Organization name
- Title of a work
- Award name

The output should be in JSON format.

Example Format Explanation:

Each entry in the JSON array consists of two fields:

1. \texttt{question}: The original input question.
2. \texttt{masked\_question}: The question with sensitive or specific details replaced by \verb|{X}|.

If the input question is empty, the \texttt{masked\_question} should also be empty.

Example:

Input:

\begin{verbatim}
[{'question': 'Who were the lead vocalists Eddie Van Halen provided
 backing vocals for in Van Halen?'},
 {'question': 'How did The Times rank Ted Hughes among British writers
  since 1945?'},
 {'question': ''},
 {'question': 'What year did Michael Crichton graduate from
  Harvard Medical School?'},
 {'question': 'In which film did Ben Affleck portray George Reeves
  and win the Volpi Cup for Best Actor?'}]
\end{verbatim}

Output:

\begin{verbatim}
[{'question': 'Who were the lead vocalists Eddie Van Halen provided
 backing vocals for in Van Halen?',
  'masked_question': 'Who were the lead vocalists {X} provided
   backing vocals for in {X}?'},
 {'question': 'How did The Times rank Ted Hughes among British
  writers since 1945?',
  'masked_question': 'How did {X} rank {X} among British writers
   since {X}?'},
 {'question': '', 'masked_question': ''},
 {'question': 'What year did Michael Crichton graduate from
  Harvard Medical School?',
  'masked_question': 'What year did {X} graduate from {X}?'},
 {'question': 'In which film did Ben Affleck portray George Reeves
  and win the Volpi Cup for Best Actor?',
  'masked_question': 'In which film did {X} portray {X} and win
   the {X} for {X}?'}]
\end{verbatim}

Your Input:

\verb|{Input}|
\end{tcolorbox}
\caption{Prompt template for masking.}
\end{figure}
\twocolumn
\onecolumn

\begin{figure}[ht]
\begin{tcolorbox}[width=0.9\textwidth, colback=white, colframe=black, boxsep=5pt, fontupper=\small]
\begin{verbatim}
<output_examples>
{
"entity": "Guy Ritchie",
"questions": [
{
"question": "Who played the lead role in Guy Ritchie's Sherlock Holmes films?",
"answers": [
"Robert Downey Jr.",
"Downey Jr."
]
},
{
"question": "Which critically acclaimed film did Guy Ritchie release in 2000?",
"answers": [
"Snatch"
]
}
]
}
</output_examples>

Create short answer questions about the entity provided, using the passage below
as a reference.

- The questions must be based on the given passage.
- The questions should be short answer questions.
- Include answer aliases in the answers field to account for variations in correct 
  responses.
- Provide the output in JSON format as shown in output_examples.
- Generate 40 questions.

Entity Name: 
{name}
Entity Passage:
{passage}
\end{verbatim}
\end{tcolorbox}
\caption{Prompt template for generating QA pairs for target and neighboring entities.}
\end{figure}
\twocolumn
\onecolumn

\begin{figure}[ht]
\begin{tcolorbox}[width=0.9\textwidth, colback=white, colframe=black, boxsep=5pt, fontupper=\small]
\begin{verbatim}
Fill in the 'x' part in the given question format to create a question. Satisfy
the following conditions.

1. It should be a question that everyone can answer.
2. If the question format is given, provide the question in JSON format.
3. The 'name' field contains information about who the question is about, 'question'
contains the question.

<output_example>
### Input:
what was the character of xs music until x
### Output:
{"name": "Michael Tippett",
"question": "What was the character of Michael Tippett's music 
            until the mid-to-late 1950s?"}
</output_example>

<entity_names_you_can_use>
{list of entity names}
</entity_names_you_can_use>

### Input:
{Input}
\end{verbatim}
\end{tcolorbox}
\caption{Prompt template for generating QA pairs for syntactically similar clusters.}
\end{figure}
\twocolumn

\section{Detailed Forget Quality and Model Utility for Each Method in Each Experiment}
\label{appendix:detailedResultsPerMethods}
\setlength{\tabcolsep}{2.5pt}
\small

\begin{table}[hbt!]
\centering


\resizebox{\columnwidth}{!}{%
\begin{tabular}{l c c c c}
\hline
 & Forget Quality & \multicolumn{3}{c}{Model Quality} \\
\cline{3-5}
 & & Entity & Domain & SynSimilar \\
\hline
Original & 0.300 & 0.712 & 0.727 & 0.770 \\
GA       & 0.734 & 0.411 & 0.415 & 0.375 \\
NPO      & 0.745 & 0.407 & 0.416 & 0.370 \\
IDK      & 0.657 & 0.199 & 0.190 & 0.174 \\
DPO      & 0.721 & 0.171 & 0.218 & 0.135 \\
\hline
\end{tabular}%
}

\caption{Forget quality and model utility for each unlearning method in a real-world scenario.}
\end{table}
\setlength{\tabcolsep}{2.5pt}
\small
\begin{table}[hbt!]
\begin{tabular}{l c c c}
\hline
 & Forget Quality & \multicolumn{2}{c}{Model Quality} \\
\cline{3-4}
 & & Domain & SynSimilar \\
\hline
Original & 0.196 & 0.973 & 0.997 \\
GA       & 0.676 & 0.565 & 0.646 \\
NPO      & 0.710 & 0.381 & 0.390 \\
IDK      & 0.685 & 0.287 & 0.357 \\
DPO      & 0.686 & 0.439 & 0.580 \\
\hline
\end{tabular}
\caption{Forget quality and model utility for each unlearning method in TOFU.}
\end{table}

\setlength{\tabcolsep}{1pt}
\small
\begin{table}[hbt!]
\centering
\resizebox{\columnwidth}{!}{%
\begin{tabular}{l c c c c}
\hline
& & \multicolumn{3}{c}{Model Utility} \\
\cline{3-5}
Method & Forget Quality & Entity & Domain & SynSimilar \\
\hline
Original & 0.300 & 0.712 & 0.727 & 0.770 \\
GA+KL    & 0.728 & 0.408 & 0.287 & 0.384 \\
GA+GD    & 0.714 & 0.317 & 0.356 & 0.239 \\
NPO+KL   & 0.689 & 0.382 & 0.423 & 0.419 \\
NPO+GD   & 0.672 & 0.345 & 0.386 & 0.409 \\
IDK+KL   & 0.653 & 0.202 & 0.194 & 0.136 \\
IDK+GD   & 0.697 & 0.166 & 0.170 & 0.152 \\
DPO+KL   & 0.717 & 0.165 & 0.198 & 0.173 \\
DPO+GD   & 0.694 & 0.389 & 0.394 & 0.373 \\
\hline
\end{tabular} %
}
\caption{Forget quality and model utility for each unlearning method with regularization using a domain neighbor set in a real-world scenario.}
\end{table}

\setlength{\tabcolsep}{2.5pt}
\small
\begin{table}[hbt!]
\centering
\resizebox{\columnwidth}{!}{%
\begin{tabular}{l c c c c}
\hline
& & \multicolumn{3}{c}{Model Utility} \\
\cline{3-5}
Method & Forget Quality & Entity & Domain & SynSimilar \\
\hline
Original & 0.300 & 0.712 & 0.727 & 0.770 \\
GA+KL    & 0.721 & 0.315 & 0.313 & 0.280 \\
GA+GD    & 0.667 & 0.432 & 0.459 & 0.287 \\
NPO+KL   & 0.679 & 0.426 & 0.453 & 0.432 \\
NPO+GD   & 0.728 & 0.413 & 0.422 & 0.371 \\
IDK+KL   & 0.687 & 0.170 & 0.169 & 0.153 \\
IDK+GD   & 0.662 & 0.201 & 0.204 & 0.136 \\
DPO+KL   & 0.687 & 0.239 & 0.169 & 0.140 \\
DPO+GD   & 0.665 & 0.373 & 0.367 & 0.407 \\
\hline
\end{tabular} %
}
\caption{Forget quality and model utility for each unlearning method with regularization using a entity neighbor set in a real-world scenario.}
\end{table}
\setlength{\tabcolsep}{2.5pt}
\small
\begin{table}
\centering
\resizebox{\columnwidth}{!}{%
\begin{tabular}{l c c c c}
\hline
& & \multicolumn{3}{c}{Model Utility} \\
\cline{3-5}
Method & Forget Quality & Entity & Domain & SynSimilar \\
\hline
Original & 0.300 & 0.712 & 0.727 & 0.770 \\
GA+KL    & 0.718 & 0.481 & 0.484 & 0.463 \\
GA+GD    & 0.657 & 0.362 & 0.395 & 0.535 \\
NPO+KL   & 0.653 & 0.443 & 0.492 & 0.491 \\
NPO+GD   & 0.729 & 0.413 & 0.424 & 0.374 \\
IDK+KL   & 0.685 & 0.178 & 0.175 & 0.161 \\
IDK+GD   & 0.655 & 0.198 & 0.225 & 0.257 \\
DPO+KL   & 0.714 & 0.215 & 0.194 & 0.168 \\
DPO+GD   & 0.658 & 0.347 & 0.405 & 0.474 \\
\hline
\end{tabular} %
}
\caption{Forget quality and model utility for each unlearning method with regularization using a syntactically similar neighbor set in a real-world scenario.}
\end{table}
\setlength{\tabcolsep}{2.5pt}
\small
\begin{table}
\centering
\begin{tabular}{l c c c}
\hline
 & Forget Quality & \multicolumn{2}{c}{Model Quality} \\
\cline{3-4}
 & & SynDifferent & SynSimilar \\
\hline
Original & 0.300 & 0.617 & 0.702 \\
GA       & 0.734 & 0.313 & 0.310 \\
NPO      & 0.745 & 0.315 & 0.297 \\
IDK      & 0.657 & 0.150 & 0.155 \\
DPO      & 0.721 & 0.127 & 0.122 \\
\hline
\end{tabular}
\caption{Forget quality and model utility for each unlearning method in a real-world scenario in paraphrasing experiments.}
\end{table}
\begin{table*}[t]
\centering
\begin{tabular}{l c c c c c c}
\hline
 & Forget Quality & \multicolumn{5}{c}{Model Quality} \\
\cline{3-7}
 & & Human & Company & Creative Works & Fictional Character & Products \\
\hline
Original & 0.300 & 0.770 & 0.623 & 0.655 & 0.575 & 0.637 \\
GA       & 0.734 & 0.375 & 0.099 & 0.108 & 0.119 & 0.110 \\
NPO      & 0.745 & 0.370 & 0.099 & 0.108 & 0.119 & 0.110 \\
IDK      & 0.657 & 0.174 & 0.085 & 0.080 & 0.080 & 0.060 \\
DPO      & 0.721 & 0.721 & 0.155 & 0.106 & 0.098 & 0.115 \\
\hline
\end{tabular}
\caption{Forget quality and model utility for each unlearning method in a real-world scenario in domain effect experiments.}
\end{table*}
\begin{table*}[h]
    \centering
    \begin{tabular}{lccccccc}
        \hline
        \textbf{Method} & \textbf{lr} & \textbf{epochs} & \textbf{Forget Efficacy} & \textbf{EntityNeigh} & \textbf{DomainNeigh} & \textbf{SynNeigh} \\
        \hline
        GA  & 5e-06  & 3 & 0.734 & 42.28 & 42.92 & \textbf{51.30} \\
        GA  & 6e-06  & 3 & 0.731 & 41.15 & 40.72 & \textbf{48.44} \\
        NPO & 4e-05  & 5 & 0.751 & 47.33 & 43.05 & \textbf{52.73} \\
        NPO & 3e-05  & 4 & 0.749 & 43.40 & 42.64 & \textbf{51.43} \\
        IDK & 2e-06  & 4 & 0.650 & 71.63 & 73.45 & \textbf{77.14} \\
        IDK & 4e-06  & 4 & 0.722 & 77.39 & 77.99 & \textbf{81.43} \\
        DPO & 6e-06  & 4 & 0.600 & 48.60 & 47.73 & \textbf{57.79} \\
        DPO & 7e-06  & 4 & 0.622 & 55.34 & 52.27 & \textbf{70.26} \\
        \hline
    \end{tabular}
    \caption{Effect of hyperparameters in the real-world scenario.}
    \label{tab:hyperparametereffects}
\end{table*}
\end{document}